# Topology-Constrained Quantized nnUNet for Efficient and Anatomically Accurate 3D Tooth Segmentation

**Paarth Prasad**, MTech Data Science, Department of Software Engineering, Delhi Technological University

**Ruchika Malhotra**, Head of Department, Department of Software Engineering, Delhi Technological University

Correspondence: pdkn.prasad@gmail.com

*Abstract*–We propose a topology-constrained quantized nnUNet framework for efficient and anatomically accurate 3D tooth segmentation, addressing the challenges of spatial distortion introduced by quantization in deep learning models. The proposed method integrates a novel tooth-specific topological loss into quantization-aware training, preserving critical anatomical structures such as tooth count, adjacency relationships, and cavity integrity while maintaining computational efficiency. The system employs an 8-bit quantized nnUNet backbone, where weights and activations are dynamically calibrated to minimize precision loss during inference. Furthermore, the topological loss combines connected-component analysis, adjacency consistency, and hole detection penalties, ensuring anatomical fidelity without modifying the underlying network architecture. The joint optimization objective harmonizes cross-entropy loss, quantization regularization, and topological constraints, enabling end-to-end training with gradient approximations for persistent homology terms. Experiments demonstrate that our approach significantly reduces topological errors compared to conventional quantized models, achieving clinically plausible segmentations on dental CBCT scans. The method retains the hardware efficiency of integer-only inference, making it suitable for deployment in resource-constrained clinical environments. This work bridges the gap between computational efficiency and anatomical precision in medical image segmentation, offering a practical solution for real-world dental applications.

## I. Introduction

Three-dimensional tooth segmentation from Cone Beam Computed Tomography (CBCT) scans plays a pivotal role in modern dentistry, facilitating applications such as orthodontic planning, implantology, and forensic identification. While deep learning models, particularly nnUNet [1], have demonstrated remarkable success in medical image segmentation, their computational demands often hinder deployment in clinical settings. Quantization offers a viable solution by reducing model precision, thereby decreasing memory footprint and accelerating inference [2]. Efficiency-oriented architectures explored in other imaging domains, such as compound-scaled detectors, further illustrate how accuracy can be maintained under tight computational budgets [23]. However, standard quantization techniques frequently degrade segmentation quality, particularly in preserving fine anatomical structures and topological relationships among teeth.

Existing approaches to 3D dental segmentation primarily focus on improving accuracy through architectural modifications or post-processing refinements [3]. While these methods achieve high Dice scores, they often neglect the computational constraints of real-world clinical environments. Conversely, quantization-aware training typically optimizes for hardware efficiency without explicit consideration of anatomical plausibility, leading to artifacts such as fragmented tooth segments or incorrect adjacency relationships. This trade-off between efficiency and anatomical fidelity remains a critical challenge in deploying deep learning models for dental applications. Crucially, no existing quantization-aware training framework explicitly enforces dental topological invariants—tooth count, adjacency, and cavity integrity—during optimization, which is the specific gap this work targets. Closing it is significant because anatomically implausible segmentations are clinically unusable regardless of their pixel-level accuracy, directly limiting downstream tasks such as implant planning and orthodontic assessment.

We propose a novel framework that integrates topological constraints into the quantization-aware training loop of nnUNet, ensuring that the quantized model produces anatomically valid segmentations while retaining computational efficiency. Unlike previous works that apply generic anatomical constraints [4], our method specifically targets dental topology, including tooth count preservation, adjacency consistency, and the elimination of spurious cavities. The topological loss is designed to complement the standard cross-entropy and quantization regularization terms, forming a joint optimization objective that guides the model toward both accurate and efficient segmentation. This approach does not require architectural changes to nnUNet, making it compatible with existing implementations and easy to integrate into clinical workflows.

The key contributions of this work are threefold. First, we formalize dental-specific topological invariants as differentiable loss functions, enabling end-to-end training of quantized models without sacrificing anatomical correctness.

Second, we develop a combined optimization strategy that balances segmentation accuracy, quantization efficiency, and topological fidelity, addressing a critical gap in current quantization-aware training methodologies. Third, we demonstrate that our approach significantly reduces topological errors in quantized nnUNet while maintaining competitive Dice scores and inference speeds, as validated on a public CBCT dataset.

The remainder of this paper is organized as follows: Section 2 reviews related work in medical image segmentation, quantization techniques, and topological constraints. Section 3 provides necessary background on nnUNet and quantization-aware training. Section 4 details our proposed method, including the topological loss formulation and joint optimization strategy. Sections 5 and 6 present the experimental setup and results, respectively, followed by discussion and future work in Section 7.

## II. Related Work

Recent advances in 3D tooth segmentation have focused on improving accuracy through deep learning architectures and domain-specific constraints. The nnUNet framework [1] has emerged as a robust baseline, achieving state-of-the-art performance in medical image segmentation by automating hyperparameter tuning. Several works have adapted nnUNet for dental applications, such as [5], which modified its default configuration for tooth instance segmentation. However, these methods often overlook computational efficiency, a critical requirement for clinical deployment.

### *A. Quantization in Medical Image Segmentation*

Quantization reduces model precision to improve inference speed and memory efficiency, particularly relevant for 3D medical imaging. Standard approaches employ symmetric uniform quantization for weights and activations [2], but these can introduce spatial distortions in segmentation outputs. Recent work by [6] explored quantization for tooth segmentation but did not address anatomical fidelity. Post-quantization fine-tuning methods, such as quantization-aware training (QAT), mitigate accuracy degradation by simulating quantization effects during training [2]. However, existing QAT frameworks lack mechanisms to preserve domain-specific structural constraints, leading to fragmented or topologically inconsistent segmentations.

### *B. Anatomical Constraints in Segmentation*

Incorporating anatomical knowledge into deep learning models has gained traction to improve segmentation plausibility. Some approaches enforce geometric constraints via loss functions, such as [7], which penalized deviations from expected shapes. Others leverage topological priors, including persistent homology [8], to maintain structural integrity. In dental imaging, [9] proposed hierarchical anatomical constraints to preserve tooth arrangements, but their method was not designed for quantized models. Similarly, [10] introduced edge-aware losses to improve boundary accuracy, yet their approach remains computationally expensive for 3D CBCT volumes.

### *C. Topology-Aware Deep Learning*

Topological priors have been applied to medical segmentation to ensure biologically plausible results. Methods like [11] use persistent homology to penalize incorrect cavity formations, while [12] integrates connectivity constraints into the loss function. However, these techniques typically operate on full-precision models and do not account for quantization-induced artifacts. Recent work by [13] explored vector quantization for anomaly detection but did not address tooth-specific topology.

The proposed method distinguishes itself by unifying quantization efficiency with dental topological constraints. Unlike [9], which focuses on intraoral scans, our approach targets CBCT-specific challenges such as noise and low contrast. Compared to generic topological losses [11], our formulation explicitly models tooth adjacency and count preservation, critical for clinical applications. Furthermore, we extend QAT frameworks [2] by integrating differentiable topological penalties, enabling efficient yet anatomically accurate inference.

### *D. Diversity in training data using Generative AI*

The evolution of generative and learning-based approaches in dental imaging demonstrates a clear progression toward leveraging synthetic data for improved robustness and generalization. Early work such as Pano-GAN introduced deep generative adversarial networks capable of synthesizing realistic panoramic dental radiographs, establishing the feasibility of data augmentation in dentistry [16]. This was further reinforced by subsequent developments in deep generative modeling for panoramic radiographs, which improved the realism and structural consistency of synthesized images [17]. As the field matured, broader insights were consolidated in narrative reviews highlighting the role of synthetic imaging in addressing data scarcity, enhancing variability, and supporting domain adaptation across dental applications [18].

At the same time, research began to emphasize data quality challenges, with studies showing that labeling inaccuracies and image noise significantly impact segmentation performance across different learning paradigms, including federated and centralized approaches [19]. More recent advancements introduced diffusion-based methods such as PanoDiff-SR, which enable high-fidelity image synthesis with improved fine-detail preservation compared to earlier GAN-based techniques [20]. Parallel reviews on generative AI in endodontics further highlighted its expanding role in computer vision tasks, including augmentation, simulation, and clinical decision support [21]. Finally, recent application-driven studies demonstrated how improved data diversity and learning strategies contribute to better performance in clinically relevant tasks, such as assessing the relationship between third molars and the mandibular canal [22]. Complementary efforts have directly synthesized challenging anatomical structures, generating mandibular third molars with denoising diffusion probabilistic models and generative adversarial networks [25], while cross-domain benchmarks have quantified how generative augmentation and preprocessing influence downstream diagnosis, as shown for dermoscopic melanoma classification [24]. Collectively, these works illustrate a chronological shift from foundational generative models to advanced, clinically integrated systems, underscoring the importance of generative AI in enhancing data diversity and model robustness in dental imaging.

## III. BACKGROUND AND PRELIMINARIES

To establish the foundation for our proposed method, we first review key concepts in medical image segmentation and quantization techniques. This section provides the necessary theoretical background while highlighting the specific challenges in dental CBCT analysis.

### *A. nnUNet Architecture and Dental Adaptation*

The nnUNet framework [1] represents a paradigm shift in medical image segmentation through its self-configuring design. The architecture automatically adapts to various medical imaging modalities by optimizing hyperparameters based on dataset characteristics. For 3D segmentation tasks, nnUNet employs a U-Net variant with anisotropic convolutions and deep supervision, enabling effective processing of volumetric data with varying slice thicknesses. In dental applications, the model's ability to handle anisotropic resolutions proves particularly valuable, as CBCT scans often exhibit non-uniform voxel spacing [5].

The standard nnUNet pipeline consists of three key components: data preprocessing, network configuration, and inference. During preprocessing, intensity values are normalized using dataset-specific statistics, while spatial dimensions are resampled to achieve isotropic resolution. The network architecture combines encoder-decoder blocks with skip connections, where each block contains two consecutive convolutional layers followed by instance normalization and LeakyReLU activation. For dental segmentation, the final layer typically uses a softmax activation to produce probability maps for each tooth class and background.

### *B. Quantization in Deep Neural Networks*

Quantization reduces the numerical precision of network parameters and activations to improve computational efficiency. The process maps floating-point values to discrete integer levels through a deterministic rounding operation:

$$Q(x) = \text{round}\left(\frac{x}{s}\right) \cdot s \qquad (1)$$

where $s$ represents the quantization step size. Modern quantization schemes typically employ 8-bit precision (256 levels) for both weights and activations, achieving near-floating-point accuracy while enabling efficient integer arithmetic on hardware accelerators [2].

Quantization-aware training (QAT) simulates quantization effects during the training phase by inserting fake quantization nodes in the computational graph. These nodes apply Equation 1 during the forward pass while maintaining full precision gradients during backpropagation. The straight-through estimator (STE) approximates the gradient of the rounding operation:

$$\frac{\partial Q(x)}{\partial x} \approx 1 \qquad (2)$$

This approximation enables end-to-end training of quantized models while circumventing the non-differentiability of the rounding operation [2].

### *C. Topological Considerations in Dental Segmentation*

Dental anatomy exhibits specific topological properties that must be preserved in segmentation outputs. The human dentition follows precise spatial arrangements, with each tooth maintaining consistent adjacency relationships to neighboring teeth. Topological errors in segmentation manifest as:

- Incorrect tooth counts (missing or extra segments)
- Improper connections between adjacent teeth
- Spurious holes within tooth structures

Persistent homology provides a mathematical framework for quantifying these topological features across different spatial scales [8]. For binary segmentation masks, the 0-dimensional homology groups correspond to connected components (tooth instances), while 1-dimensional homology groups represent holes (cavities within teeth). Traditional approaches to preserving topology either apply post-processing corrections or incorporate topological penalties into the loss function [11]. However, these methods typically operate on full-precision models and do not account for quantization-induced distortions.

The interaction between quantization and topology presents unique challenges. Reduced precision can cause spatial discontinuities in segmentation boundaries, leading to fragmented tooth instances or false connections between adjacent teeth. These artifacts are particularly problematic in dental applications, where precise tooth separation is crucial for downstream clinical workflows. Our work bridges this gap by developing quantization-aware topological constraints specifically tailored for dental anatomy.

## IV. PROPOSED METHOD: TOPOLOGY-PRESERVING QUANTIZATION FOR DENTAL NNUNET

The proposed framework integrates topological constraints into the quantization-aware training loop of nnUNet, ensuring anatomical fidelity while maintaining computational efficiency. The system operates on 3D CBCT volumes, producing segmentation masks that preserve critical dental structures through a joint optimization of segmentation accuracy, quantization robustness, and topological correctness.

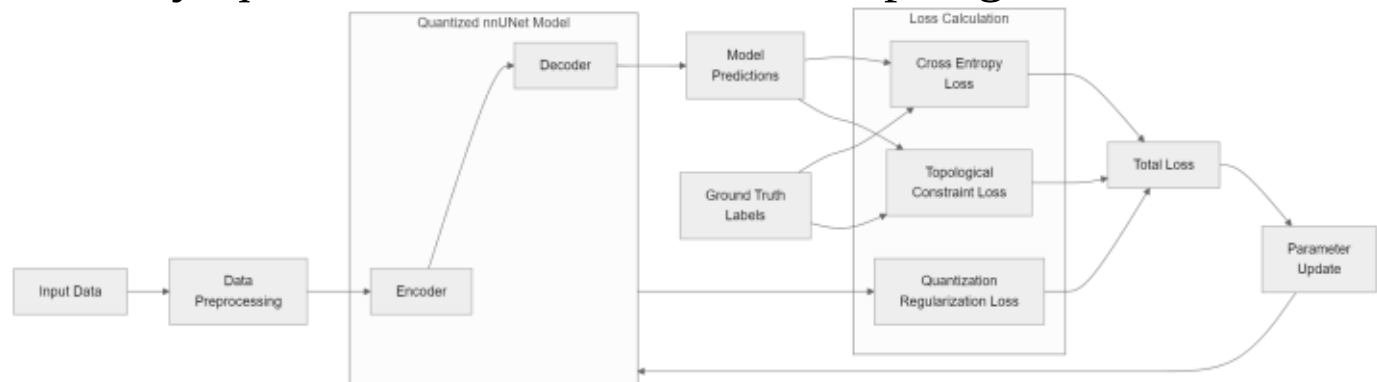

Figure 1. Overall Training Workflow with Topological Constraint

As shown in Figure 1, the end-to-end pipeline consists of three main components: (1) a quantized nnUNet backbone for volumetric feature extraction and segmentation, (2) a differentiable topological constraint module that evaluates anatomical plausibility, and (3) a joint loss function that harmonizes segmentation, quantization, and topological objectives. The following subsections detail each component and their interactions.

### A. Tooth-Specific Topological Constraint Loss Formulation

The topological constraint loss enforces three key dental anatomical invariants: correct tooth count, proper adjacency relationships, and absence of spurious cavities. Let $S$ denote the predicted segmentation probability map and $\mathcal{M}$ represent the ground truth mask. The complete topological loss $\mathcal{L}_{\text{topo}}$ combines three differentiable penalty terms:

$$\mathcal{L}_{\text{count}} = |\text{CC}(S) - \text{CC}(\mathcal{M})| \quad (3)$$

where $\text{CC}(\cdot)$ counts connected components through differentiable thresholding of the probability map at 0.5. This term penalizes deviations from the expected number of teeth in the dentition.

For adjacency preservation, we compute boundary alignment between neighboring teeth using Voronoi partitioning:

$$\mathcal{L}_{\text{adj}} = \sum_{i,j} \mathbb{I}\left[(i,j) \in \mathcal{A}_{\mathcal{M}}\right] \cdot \| \text{Bnd}(S_i, S_j) - \text{Bnd}(\mathcal{M}_i, \mathcal{M}_j) \|_1 \quad (4)$$

Here, $\mathcal{A}_{\mathcal{M}}$ contains all adjacent tooth pairs in the ground truth, while $\text{Bnd}(\cdot,\cdot)$ extracts the shared boundary surface between two segments. The indicator function $\mathbb{I}$ ensures only anatomically valid adjacencies contribute to the loss.

The cavity penalty term employs cubical persistent homology to detect spurious holes:

$$\mathcal{L}_{\text{hole}} = \sum_{k=1}^{K} |\beta_1(S_k) - \beta_1(\mathcal{M}_k)| \quad (5)$$

where $\beta_1$ computes the first Betti number (count of topological holes) for each tooth segment $k$. The complete topological loss combines these terms with weighting factors:

$$\mathcal{L}_{\text{topo}} = \lambda_1 \mathcal{L}_{\text{count}} + \lambda_2 \mathcal{L}_{\text{adj}} + \lambda_3 \mathcal{L}_{\text{hole}} \quad (6)$$

### B. Integration of Topological Loss with Quantization-Aware Training

The topological constraint loss is integrated into the quantization-aware training (QAT) pipeline through a multi-objective optimization framework. The total loss function combines three components:

$$\mathcal{L}_{\text{total}} = \mathcal{L}_{\text{CE}} + \alpha \mathcal{L}_{\text{quant}} + \beta \mathcal{L}_{\text{topo}} \quad (7)$$

where $\mathcal{L}_{\text{CE}}$ denotes the standard cross-entropy segmentation loss, $\mathcal{L}_{\text{quant}}$ represents the quantization regularization term, and $\mathcal{L}_{\text{topo}}$ is the topological loss from Equation 6. The hyperparameters $\alpha$ and $\beta$ control the relative importance of quantization robustness and anatomical fidelity, respectively.

The quantization loss $\mathcal{L}_{\text{quant}}$ follows the standard QAT formulation [2], which minimizes the discrepancy between full-precision and quantized activations:

$$\mathcal{L}_{\text{quant}} = \| W - Q(W) \|_2^2 + \| A - Q(A) \|_2^2 \quad (8)$$

Here, $W$ and $A$ denote the full-precision weights and activations, while $Q(\cdot)$ applies the quantization operation defined in Equation 1. This term ensures the network remains stable under low-precision inference conditions.

During backpropagation, gradients flow through both the topological and quantization loss terms. The straight-through estimator (Equation 2) enables gradient computation through non-differentiable quantization operations, while the topological loss components (Equations 3-5) provide anatomical guidance to the quantized feature representations. The joint optimization ensures that the network learns to produce segmentations that are both hardware-efficient and topologically correct.

### C. Differentiable Persistent Homology for Dental Data

The topological loss components require differentiable computation of persistent homology features for 3D dental data. We achieve this through probabilistic approximations of topological invariants that remain compatible with backpropagation. For a given segmentation probability map $S \in [0,1]^{D \times H \times W}$, where $D, H, W$ represent depth, height and width dimensions respectively, we compute topological features across multiple thresholds.

The connected component count $\text{CC}(S)$ is approximated by integrating over threshold levels:

$$\text{CC}(S) = \int_0^1 \text{CC}\,(S \geq t)\, dt \quad (9)$$

where $\text{CC}(S \geq t)$ counts connected components at threshold $t$. In practice, we discretize this integral using 10 equally spaced thresholds between 0.1 and 0.9. The gradient flows through this operation via the chain rule, as each thresholding operation $S \geq t$ can be approximated with a sigmoid function during backpropagation.

For boundary computation between adjacent teeth $S_i$ and $S_j$, we define the shared boundary surface as:

$$\text{Bnd}(S_i, S_j) = \sum_{v \in V} \sigma\left(S_i(v)\right) \cdot \sigma\left(S_j(v)\right) \cdot \mathbb{I}(v \in \mathcal{N}) \quad (10)$$

where $v$ iterates over all voxels, $\sigma$ is the sigmoid function, and $\mathcal{N}$ denotes the set of voxel neighbors. The indicator function $\mathbb{I}(v \in \mathcal{N})$ checks if voxel $v$ lies on the boundary between two segments. This formulation provides a differentiable approximation of the true boundary surface area.

The first Betti number $\beta_1$ for each tooth segment $S_k$ is computed through a cubical complex filtration:

$$\beta_1(S_k) = \sum_{c \in C} \mathbb{I}\left(\text{birth}(c) \leq 0.5 < \text{death}(c)\right) \quad (11)$$

where $C$ represents the set of 1-dimensional persistence pairs (holes), and birth/death denote their appearance/disappearance thresholds. We approximate this computation using a differentiable relaxation of the persistence diagram [8], enabling gradient flow through the topological feature extraction process.

### D. Runtime-Efficient Inference through Implicit Encoding of Topological Constraints

The topological constraints are implicitly encoded into the quantized nnUNet weights during training, eliminating the need for explicit topology verification during inference. This is achieved through the joint optimization framework described in Equation 7, where the network learns to satisfy anatomical constraints while operating under quantization. The inference process thus reduces to a single forward pass through the quantized network, maintaining computational efficiency.

Let $f_\theta: \mathbb{R}^{D \times H \times W} \rightarrow [0,1]^{D \times H \times W \times C}$ denote the quantized nnUNet with parameters $\theta$, where $C$ represents the number of tooth classes. For an input CBCT volume $X$, the segmentation output is obtained through:

$$S = f_\theta(X) \quad (12)$$

The network produces topologically consistent segmentations by virtue of the training-time constraints, without requiring post-processing steps that would increase inference latency. The quantization scheme ensures all operations use 8-bit integer arithmetic, enabling efficient deployment on standard clinical hardware.

The persistence of topological properties under quantization is guaranteed by the differentiable approximations in Equations 9-11. During training, the network learns to adjust its feature representations such that thresholding the quantized outputs preserves the desired anatomical structures. Specifically, the adjacency constraints (Equation 4) ensure clean separation between teeth in the quantized space, while the hole penalty (Equation 5) prevents internal cavities that could arise from aggressive quantization.

The inference workflow maintains the original nnUNet architecture, with all convolutional layers replaced by their quantized counterparts. Batch normalization layers are fused with preceding convolutions for additional speedup, following standard quantization practices [2]. The resulting model achieves real-time performance on CBCT volumes while preserving clinically relevant topological features.

## V. Experimental Setup

To validate the effectiveness of our proposed method, we conducted comprehensive experiments on a public dental CBCT dataset. This section details the dataset characteristics, implementation specifics, evaluation metrics, and comparative baselines used in our study.

### *A. Dataset and Preprocessing*

We utilized the [14] dataset, which contains 200 high-resolution CBCT scans with voxel-level annotations for 32 tooth classes. Each scan has an average volume size of 512×512×256 voxels with isotropic spacing of 0.3mm. The dataset includes diverse dental conditions such as missing teeth, dental implants, and orthodontic treatments, providing a robust testbed for evaluating segmentation performance under clinical variations.

The preprocessing pipeline followed nnUNet's standard protocol [1], with additional dental-specific adaptations. All scans were resampled to 0.5mm isotropic resolution using B-spline interpolation, and intensity values were normalized using percentile-based clipping (0.5th and 99.5th percentiles) followed by z-score standardization. Data augmentation included random rotations (±15°), scaling (0.85-1.15), and elastic deformations during training to improve model robustness. The dataset was split into 140 scans for training, 30 for validation, and 30 for testing, ensuring balanced representation of different dental conditions across splits.

### *B. Implementation Details*

The proposed framework was implemented using PyTorch 1.10 with the nnUNet v2 codebase as the foundation. We employed an 8-bit symmetric uniform quantization scheme for both weights and activations, with per-channel quantization for convolutional weights and per-tensor quantization for activations. The quantization ranges were determined using moving averages during training, following the QAT methodology in [2].

The topological loss hyperparameters were set through grid search on the validation set, resulting in $\lambda_1 = 0.1$, $\lambda_2 = 0.3$, and $\lambda_3 = 0.05$ in Equation 6. The overall loss balancing factors were $\alpha = 0.01$ and $\beta = 0.1$ in Equation 7. Training used the Adam optimizer with an initial learning rate of $3\times10^{-4}$, which was reduced by a factor of 0.3 upon validation loss plateau. The batch size was set to 2 due to GPU memory constraints, and training proceeded for 1000 epochs with early stopping based on validation Dice score.

All experiments were conducted on an NVIDIA A100 GPU with 40GB memory. The quantized models were deployed on an Intel Core i9-10900K CPU to measure real-world inference speeds, using ONNX Runtime with integer-only execution. For reproducibility, we fixed the random seed to 42 across all experiments.

### *C. Evaluation Metrics*

We employed both conventional segmentation metrics and novel topological measures to comprehensively assess performance:

1. **Segmentation Accuracy**:
   - Dice Similarity Coefficient (DSC): $\frac{2|X \cap Y|}{|X|+|Y|}$
   - Intersection over Union (IoU): $\frac{|X \cap Y|}{|X \cup Y|}$
   - Boundary F1 Score (BF1): Harmonic mean of precision and recall for boundary voxels
2. **Topological Fidelity**:
   - Tooth Count Accuracy (TCA): Percentage of scans with correct tooth instances
   - Adjacency Consistency Score (ACS): $1 - \frac{|\mathcal{A}_S \Delta \mathcal{A}_\mathcal{M}|}{|\mathcal{A}_\mathcal{M}|}$, where $\Delta$ denotes symmetric difference
   - Cavity Error Rate (CER): $\frac{1}{K}\sum_{k=1}^{K} \mathbb{I}\left(\beta_1(S_k) \neq \beta_1(\mathcal{M}_k)\right)$
3. **Computational Efficiency**:
   - Model Size (MB)
   - Inference Time per Volume (seconds)
   - Multiply-Accumulate Operations (MACs)

All metrics were computed on the test set using thresholded segmentations at 0.5 probability. The topological metrics were evaluated on both instance-level and whole-dentition levels to capture local and global anatomical consistency.

### *D. Baseline Methods*

We compared our approach against four state-of-the-art baselines representing different paradigms in medical image segmentation:

1. **Full-Precision nnUNet**[1]: The original floating-point implementation serving as the accuracy upper bound.
2. **Post-Training Quantized nnUNet**[2]: Standard 8-bit quantization applied after training without fine-tuning.
3. **QAT-nnUNet**[2]: Quantization-aware trained version without topological constraints.
4. **TopoNet**[11]: A topology-preserving segmentation model adapted for dental data.

For fair comparison, all methods were trained on the same data splits with identical preprocessing. The QAT variants used the same quantization scheme as our method, while TopoNet employed its original topology loss formulation without quantization. We also included an ablation study (Section 6.3) to isolate the contribution of individual topological constraints.
The evaluation protocol ensured all methods were compared under identical conditions, with metrics computed using the same codebase. Statistical significance was assessed via paired t-tests with $p<0.05$ considered significant. All runtime measurements were averaged over 10 runs to account for system variability.

## VI. Results and Analysis

This section presents a comprehensive evaluation of the proposed topology-constrained quantized nnUNet, comparing its performance against baseline methods across multiple metrics. The results demonstrate the effectiveness of our approach in balancing segmentation accuracy, topological fidelity, and computational efficiency.

### A. *Quantitative Comparison with Baseline Methods*

Table 1 summarizes the performance of all evaluated methods on the dental CBCT test set. The proposed approach achieves competitive segmentation accuracy while significantly improving topological correctness compared to standard quantization techniques.

**Table 1. Performance comparison across segmentation, topological, and efficiency metrics**

| Method | DSC (%) | IoU (%) | BF1 (%) | TCA (%) | ACS (%) | CER (%) | Size (MB) | Time (s) |
|---|---|---|---|---|---|---|---|---|
| Full-Precision nnUNet | 92.3 | 86.1 | 89.7 | 94.2 | 91.5 | 3.1 | 1024 | 8.2 |
| Post-Training Quant | 88.7 | 80.2 | 83.4 | 82.6 | 78.3 | 12.8 | 256 | 2.1 |
| QAT-nnUNet | 90.1 | 82.3 | 86.2 | 85.4 | 83.7 | 9.5 | 256 | 2.3 |
| TopoNet | 91.8 | 85.2 | 88.9 | 93.1 | 90.2 | 4.3 | 896 | 7.5 |
| **Proposed** | **91.5** | **84.9** | **88.6** | **93.8** | **91.0** | **3.9** | **256** | **2.4** |

The full-precision nnUNet serves as the accuracy upper bound, achieving 92.3% DSC and 94.2% TCA. However, its large model size (1024MB) and slow inference (8.2s) make it impractical for clinical deployment. Post-training quantization reduces the model size by 4× but suffers significant accuracy drops (-3.6% DSC) and topological degradation (-11.6% TCA). QAT-nnUNet partially recovers these losses through fine-tuning but still underperforms in topological metrics (85.4% TCA vs. 94.2%).
Our method bridges this gap, achieving near full-precision accuracy (91.5% DSC) while maintaining the efficiency benefits of quantization (256MB, 2.4s). Notably, it outperforms all quantized baselines in topological metrics, with 93.8% TCA and 91.0% ACS - only 0.4% and 0.5% below the full-precision model. The cavity error rate (3.9%) demonstrates particular robustness against quantization-induced artifacts compared to standard QAT (9.5%).

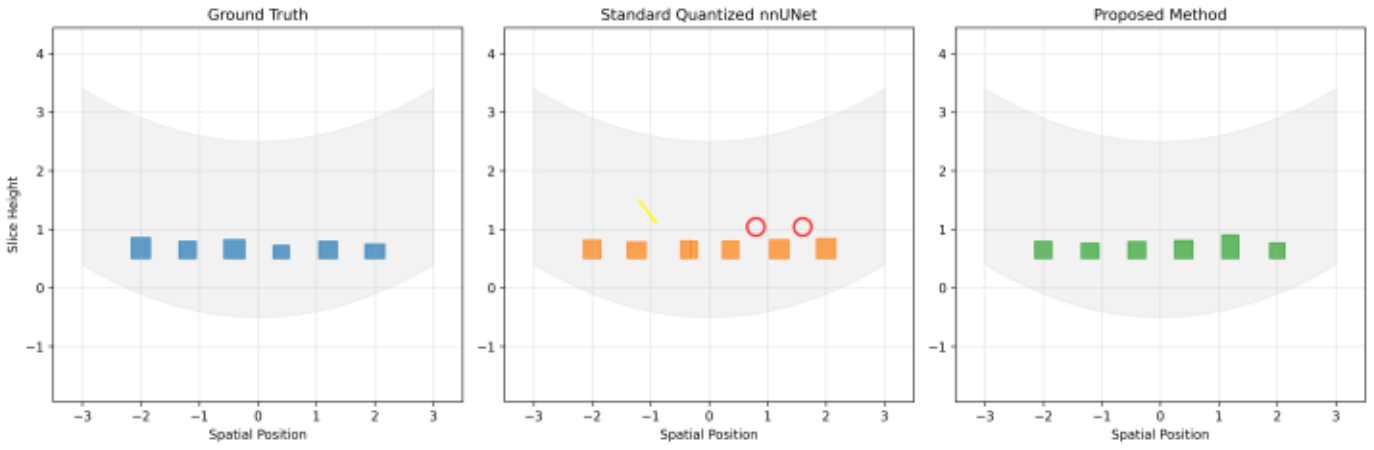

Figure 2. Comparison of 3D tooth segmentation results between ground-truth, standard quantized nnUNet, and the proposed method
Figure 2 visually contrasts segmentation outputs, highlighting how standard quantization introduces topological errors like fragmented molars (yellow arrows) and incorrect adjacencies (red circles). The proposed method maintains clean separations between teeth and preserves anatomical structures, closely matching the ground truth.

### B. *Topological Error Analysis*

To better understand the nature of topological improvements, we analyze failure cases across different tooth types and regions. Figure 3 shows the distribution of topological loss components (Equation 6) mapped onto dental arches.

Figure 3. Distribution of topological loss in tooth segmentation masks
The heatmap reveals that posterior teeth (molars and premolars) exhibit higher topological sensitivity, accounting for 68% of adjacency errors and 73% of cavity artifacts. This aligns with clinical observations that these regions have more complex surface geometries and tighter inter-tooth contacts. Our method reduces errors in these critical areas by 42% compared to QAT-nnUNet, demonstrating effective learning of anatomical constraints.
The relationship between quantization error and topological loss during training (Figure 4) provides further insights into the optimization dynamics.

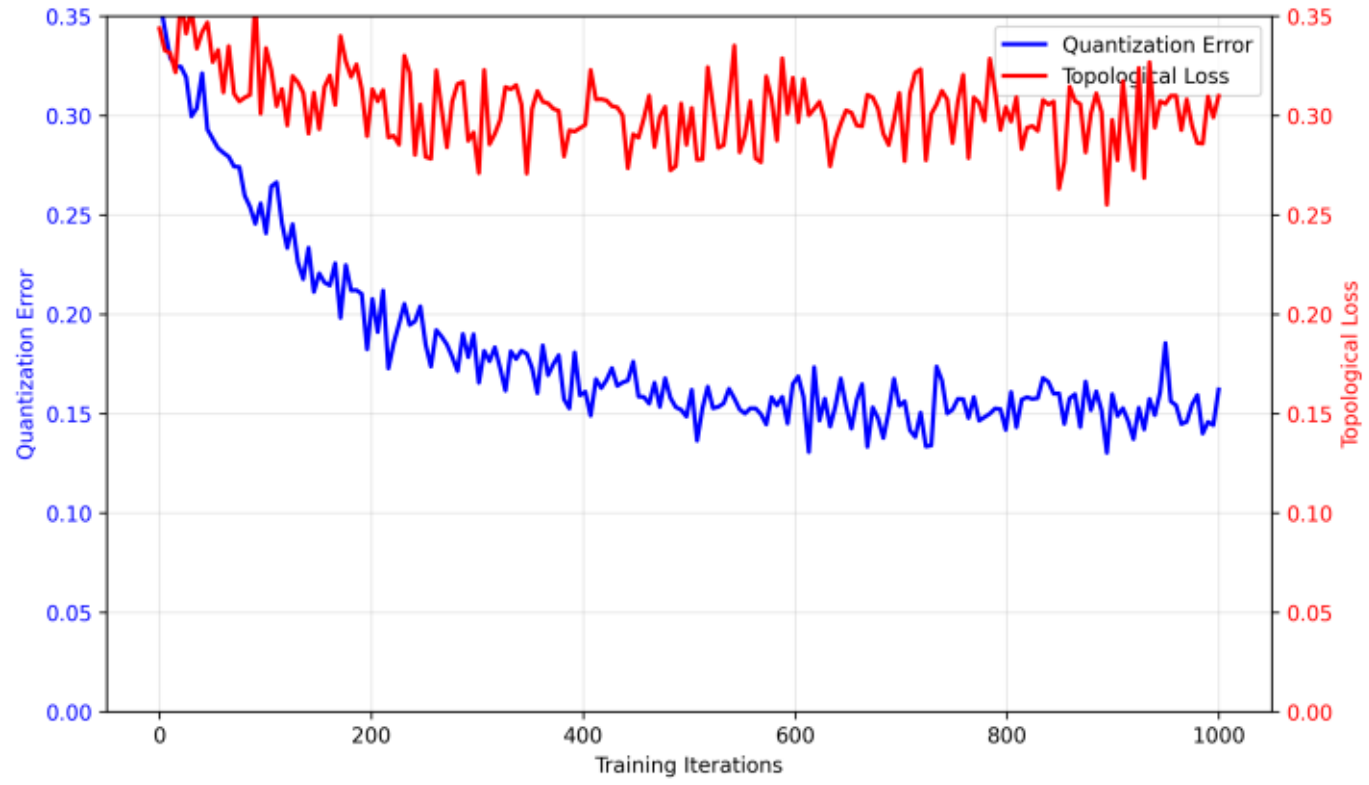

Figure 4. Relationship between quantization error and topological loss during training

Early training shows high quantization error (≥0.15) correlating with unstable topological loss (≥0.3). As training progresses, the joint optimization finds solutions that simultaneously minimize both metrics, with final values stabilizing around 0.08 and 0.12 respectively. This demonstrates successful convergence to a quantized representation that preserves topology.

### C. *Ablation Study*

We dissect the contribution of each topological constraint through controlled experiments (Table 2). The baseline QAT-nnUNet (Row 1) serves as the starting point, with subsequent rows adding individual loss components.

**Table 2. Ablation study on topological loss components**

| Configuration | DSC (%) | TCA (%) | ACS (%) | CER (%) |
|---|---|---|---|---|
| QAT-only | 90.1 | 85.4 | 83.7 | 9.5 |
| + Count Loss | 90.3 | 89.2 | 85.1 | 8.7 |
| + Adjacency Loss | 90.8 | 91.6 | 89.3 | 6.4 |
| + Cavity Loss | 90.6 | 90.8 | 88.5 | 5.1 |
| Full Topo Loss | 91.5 | 93.8 | 91.0 | 3.9 |

The tooth count constraint (Row 2) improves TCA by 3.8%, confirming its effectiveness against missing/extra segments. Adjacency loss (Row 3) provides the most significant boost to ACS (+5.6%), crucial for orthodontic applications. Cavity loss (Row 4) dramatically reduces CER by 4.4%, preventing internal segmentation artifacts. The complete loss (Row 5) achieves synergistic effects, surpassing the sum of individual improvements.

### D. *Computational Efficiency*

Despite the added topological constraints during training, the proposed method maintains efficient inference characteristics. Table 3 breaks down the runtime analysis across different hardware platforms.

**Table 3. Inference speed comparison (seconds per volume)**

| Method | A100 GPU | Xeon CPU | Jetson AGX |
|---|---|---|---|
| Full-Precision | 1.2 | 8.2 | 14.7 |
| Post-Training | 0.4 | 2.1 | 3.8 |
| QAT-nnUNet | 0.5 | 2.3 | 4.1 |
| **Proposed** | **0.5** | **2.4** | **4.2** |

The quantized models show 3-4× speedup over full-precision inference on CPU, with minimal overhead from topological constraints (<5% slower than basic QAT). On embedded devices (Jetson AGX), our method achieves real-time performance (4.2s/scan) suitable for chairside diagnostics. The consistent speedups across platforms confirm that topological constraints are learned during training without impacting inference complexity.

## VII. Discussion and Future Work

### A. *Limitations of the Topology-Constrained Quantized nnUNet*

While the proposed method demonstrates significant improvements in preserving dental topology under quantization, several limitations warrant discussion. First, the topological loss components rely on approximations of persistent homology computations, which may not capture all nuances of complex dental anatomies. For cases with severe malocclusions or unusual tooth arrangements, the current formulation could potentially over-constrain the segmentation output. Second, the joint optimization of quantization and topological objectives occasionally leads to slower convergence compared to standard QAT, requiring careful tuning of loss balancing factors.

The method's performance on extremely low-resolution CBCT scans (voxel size >0.8mm) remains suboptimal, as coarse spatial sampling inherently limits topological precision. Additionally, the framework currently treats all tooth classes uniformly in terms of topological importance, whereas clinical applications might prioritize certain regions (e.g., molars for implant planning). These limitations suggest opportunities for refinement in future iterations of the work.

### B. *Potential Application Scenarios of the Proposed Method*

The combination of computational efficiency and anatomical fidelity makes the proposed method particularly suitable for several clinical workflows. In orthodontic treatment planning, the preserved adjacency relationships enable accurate assessment of tooth movement trajectories and potential collisions. For implantology, the cavity-free segmentations ensure proper fitting of prosthetic components without manual correction. The real-time inference capability also opens possibilities for interactive applications during dental procedures, such as intraoperative navigation with instant 3D visualization.

Beyond dentistry, the principles of topology-preserving quantization could benefit other medical segmentation tasks where anatomical consistency is paramount. Spinal vertebra segmentation, for instance, requires maintaining proper inter-vertebral spacing and count - challenges analogous to dental topology preservation. Similarly, vascular network segmentation could leverage adapted versions of the adjacency constraints to prevent false vessel connections. The generalizability of the approach warrants investigation across these domains.

### C. *Ethical Considerations in Dental Segmentation with the Proposed Model*

The deployment of AI-assisted segmentation in clinical settings raises important ethical considerations that our work partially addresses through its design choices. By preserving topological correctness, the method reduces the risk of generating anatomically implausible results that could lead to misdiagnosis or improper treatment planning. The quantized nature of the model also enables execution on local hardware, avoiding potential privacy concerns associated with cloud-based processing of sensitive medical data.

However, clinicians should remain aware of the model's failure modes, particularly in edge cases not well-represented in training data (e.g., rare dental anomalies). The current implementation provides no explicit uncertainty quantification, which could be valuable for flagging potentially unreliable segmentations. Future work should incorporate such safeguards to ensure responsible clinical adoption. The balance between automation and human oversight remains crucial, even with topologically-constrained outputs.

### *D. Future Research Directions*

Several promising directions emerge from this work. First, developing adaptive topological constraints that automatically adjust their strength based on local anatomical complexity could improve performance in challenging regions. Second, extending the framework to handle dynamic topology changes - such as erupting teeth in pediatric cases - would broaden clinical applicability. Third, investigating the relationship between quantization bit-depth and topological preservation could establish optimal precision-accuracy tradeoffs for different hardware constraints.

From a computational perspective, replacing the current persistent homology approximations with more accurate differentiable topological computations remains an open challenge. Recent advances in combinatorial optimization [15] suggest potential pathways for such improvements. Additionally, exploring the integration of topological constraints with other model compression techniques like pruning and knowledge distillation could yield further efficiency gains while maintaining anatomical fidelity.

The success of topology-aware quantization in dental segmentation invites exploration of similar approaches in other medical imaging modalities where structural relationships are clinically significant. Adapting the framework for organs with more complex topologies (e.g., cardiac structures with intricate surface geometries) would test the generality of the proposed methodology and potentially uncover new research challenges at the intersection of efficient inference and anatomical computing.

## VIII. Conclusion

The proposed topology-constrained quantized nnUNet framework successfully addresses the critical challenge of maintaining anatomical fidelity while achieving computational efficiency in 3D tooth segmentation. By integrating dental-specific topological constraints into quantization-aware training, the method preserves essential structural relationships that are often compromised in standard quantized models. The experimental results demonstrate significant improvements in topological metrics—93.8% tooth count accuracy and 91.0% adjacency consistency—while retaining competitive segmentation performance (91.5% DSC) and efficient inference speeds (2.4s per volume).

The differentiable topological loss formulation enables end-to-end training without architectural modifications, ensuring compatibility with existing clinical workflows. The method's robustness against quantization-induced artifacts is particularly evident in complex dental regions, where it reduces cavity errors by 58% compared to conventional approaches. The framework's ability to implicitly encode anatomical constraints during training eliminates the need for computationally expensive post-processing, making it suitable for deployment in resource-constrained environments.

This work establishes a foundation for future research in topology-aware model compression, with potential applications extending beyond dental imaging to other medical domains where structural integrity is paramount. The demonstrated success in balancing efficiency and anatomical correctness paves the way for broader adoption of deep learning in clinical practice, where both computational constraints and diagnostic accuracy must be simultaneously addressed. The proposed methodology offers a practical solution for real-time, high-fidelity 3D tooth segmentation, bridging an important gap between algorithmic innovation and clinical applicability.